\documentclass{article}

\usepackage{PRIMEarxiv}

\usepackage[utf8]{inputenc} 
\usepackage[T1]{fontenc}    
\usepackage{hyperref}       
\usepackage{url}            
\usepackage{booktabs}       
\usepackage{amsfonts}       
\usepackage{nicefrac}       
\usepackage{microtype}      
\usepackage{lipsum}
\usepackage{fancyhdr}       
\usepackage{graphicx}       
\graphicspath{{media/}}     

\usepackage{algorithm}
\usepackage{algpseudocode}

\usepackage{subcaption}

\usepackage{fancyhdr}
\usepackage[utf8]{inputenc} 
\usepackage{hyperref}       
\usepackage{url}            
\usepackage{booktabs}       
\usepackage{amsfonts}       
\usepackage{nicefrac}       
\usepackage{microtype}      
\usepackage{lipsum}
\usepackage{fancyhdr}       
\usepackage{graphicx}       
\usepackage[numbers]{natbib}
\usepackage{amsmath}
\usepackage{subcaption}
\usepackage{tabularx}
\usepackage{booktabs}
\usepackage{svg}
\usepackage{threeparttable}

\begin{document}
\fancyhead[LO]{\textsc{LDFaceNet}}

\title{\textsc{LDFaceNet}: Latent Diffusion-based Network for High-Fidelity Deepfake Generation

}

\author{
  Dwij Mehta$^*$, Aditya Mehta$^*$, Pratik Narang \\
  Department of CSIS \\
  Birla Institute of Technology and Science, Pilani Campus \\
  Pilani, Rajasthan, India \\
  \texttt{\{f20190122,p20230303,pratik.narang\}@pilani.bits-pilani.ac.in} \\
}

\def\thefootnote{*}\footnotetext{These authors contributed equally to this work}\def\thefootnote{\arabic{footnote}}
\maketitle

\begin{abstract}
Over the past decade, there has been tremendous progress in the domain of synthetic media generation. This is mainly due to the powerful methods based on generative adversarial networks (GANs). Very recently, diffusion probabilistic models, which are inspired by non-equilibrium thermodynamics, have taken the spotlight. In the realm of image generation, diffusion models (DMs) have exhibited remarkable proficiency in producing both realistic and heterogeneous imagery through their stochastic sampling procedure. This paper proposes a novel facial swapping module, termed as \textsc{LDFaceNet} \textit{(Latent Diffusion based Face Swapping Network)},  which is based on a guided latent diffusion model that utilizes facial segmentation and facial recognition modules for a conditioned denoising process. The model employs a unique loss function to offer directional guidance to the diffusion process. Notably, \textsc{LDFaceNet} can incorporate supplementary facial guidance for desired outcomes without any retraining. To the best of our knowledge, this represents the first application of the latent diffusion model in the face-swapping task without prior training. The results of this study demonstrate that the proposed method can generate extremely realistic and coherent images by leveraging the potential of the diffusion model for facial swapping, thereby yielding superior visual outcomes and greater diversity.
\end{abstract}


\section{Introduction}
    Recently, deep learning models of all types have started producing high-quality synthetic media. This media can be visual, audio or video files. Stunning image and audio samples have been created using GANs, autoregressive models, flows, and variational autoencoders (VAEs) \cite{goodfellow2014generative,brock2018large,karras2018progressive,PrengerWaveGlow,pmlr-v80-kalchbrenner18a,RazaviVAE}. Recent advances in fields like energy-based modeling and score matching have also started producing synthetic media that are comparable to those of GANs \cite{EnergySong01}.
    
    \begin{figure}[t]
      \centering
      \includegraphics[width=12cm]{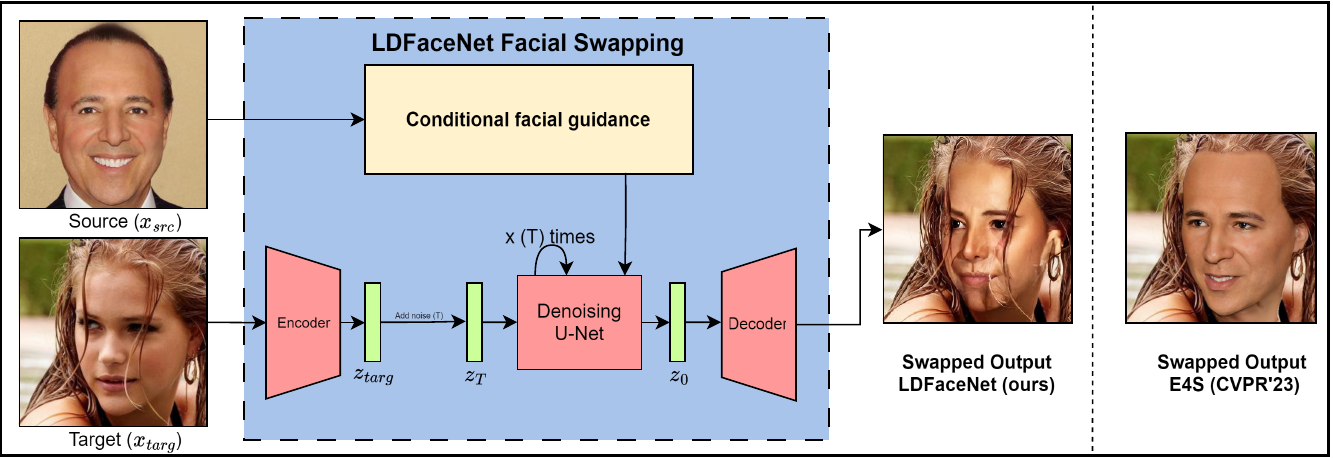}
      \caption{\textbf{Sample output of \textsc{LDFaceNet}.} Compared to recent state-of-the-art methods such as E4S (CVPR'23 \cite{liu2023e4s}), the results produced by \textsc{LDFaceNet} are significantly better. This particular example also illustrates that our method performs much better in handling occlusions over the target face than other generative methods. Further details are in the results section \ref{sec:results}.}
      \label{fig:intro}
    \end{figure}

    GANs have emerged as the state-of-the-art method for image generation tasks. The performance for generative tasks is very subjective and varies from person to person. However, there are two commonly used distribution-based sample quality metrics such as FID \cite{heusel2017gansFID} and Inception score (IS) \cite{salimans2016improvedInceptionDist} that most papers have used to report their results. Recently, GANs have been criticized for their limited diversity capture capabilities, and it has been demonstrated that likelihood-based models outperform GANs in this regard \cite{RazaviVAE}. Additionally, GANs are often difficult to train, and we need to fine-tune their hyperparameters and regularizers to avoid collapse during training. Despite the drawbacks, GANs are still considered the leading method for image generation, but they are still unable to scale and apply to new domains. Consequently, there have been efforts to achieve state-of-the-art sample quality with likelihood-based models, which offer better scalability and ease of training. However, these models still lag behind GANs in terms of visual sample quality, and their sampling process is costlier and slower than that of GANs.

    A class of likelihood-based models known as diffusion models \cite{ho2020denoising,nichol2021improved} has recently been shown to produce visually realistic images while offering desirable characteristics such as variety, a stationary training objective, and simple scalability. These models generate samples by gradually eliminating noise from a signal, and their training objective can be described as a re-weighted variational lower bound. Compared to GANs, diffusion models enable more stable training and yield more desirable results in terms of fidelity and diversity. To manage the trade-off between fidelity and diversity, classifier guidance \cite{dhariwal2021diffusion} is used to guide the diffusion process.

    In the domain of image generation, face swapping is a computer vision task that involves transferring the face of one individual (the source) to another (the target) while preserving the target's facial attributes, such as identity, expression, and pose. This task has various applications in the entertainment industry, particularly in films, where it is used to replace the face of an actor with that of a stunt double or to resurrect deceased actors. Face swapping, also widely known as deepfakes generation can also be used for practical purposes, such as in forensic investigations and for facial reconstruction in the medical domain. 

    In this paper, we introduce a novel guided diffusion model, \textsc{LDFaceNet}, for deepfake generation. To the best of our knowledge, no prior research has explored face swapping using pre-trained latent diffusion models. Training diffusion models from scratch demands extensive computational resources and careful hyperparameter tuning. Our method, however, eliminates the need for re-training by leveraging the weights provided by Rombach et al. \cite{rombach2022high} from their LDM trained on the CelebA dataset \cite{liu2018largeCelebA}. We enhance this LDM with a unique facial guidance module. By using embeddings of images generated during intermediate timesteps, our model is constrained and guided through the facial guidance module. Additionally, we implement latent-level blending to ensure a seamless transition at the boundaries of the swapped face. This approach not only proves to be cost-effective but also outperforms existing facial swapping methods in both qualitative and quantitative evaluations by great margins. Furthermore, our method demonstrates robustness in handling faces with occlusions, misalignments, or non-frontal views, making it highly versatile in various challenging scenarios.

\section{Related Work}

\subsection{Models for Image Synthesis}
    \subsubsection*{GANs}    
    Generative modeling faces unique challenges due to the large size of modern-day images. GANs \cite{goodfellow2014generative} enable the effective synthesis of visually realistic images with good perceptual quality \cite{brock2018large}, but they are difficult to optimize and struggle to capture the complete data distribution. While likelihood-based methods prioritize accurate density estimation, their optimization behaves more reliably. Variational autoencoders (VAEs) and flow-based models can synthesize high-resolution pictures effectively, but their sample quality is generally inferior to that of GANs \cite{vahdat2020nvae}. Autoregressive models (ARMs) \cite{van2016pixel, child2019generating}, despite their good performance in density estimation, are limited by their sequential sampling procedure and computationally expensive designs \cite{vaswani2017attention}, which restrict the resolution of the images they can produce. Maximum-likelihood training expends a disproportionate amount of capacity to model the scarcely perceptible, high-frequency details present at the pixel level, leading to lengthy training durations. To address this, several two-stage approaches first compress an image to a latent image space using ARMs rather than processing raw pixels, allowing for scaling to higher resolutions.

    \subsubsection*{Diffusion Probabilistic Models}
    Recently, Diffusion Probabilistic Models (DM)\cite{ho2020denoising} have produced cutting-edge outcomes in sample quality. When their learned posterior or learned network's backbone is applied as a U-Net, these models are a natural fit for image-like data.
    
    
    However, the disadvantage of evaluating and optimizing these models in pixel space is low inference speed because of repeated sequential sampling and very high training costs. While the former can be addressed in part by sophisticated sampling techniques like implicit diffusion Models \cite{song2020denoising} and hierarchical approaches \cite{vahdat2021score}, training on high-resolution image data always necessitates the calculation of expensive gradients.  Latent diffusion models (LDMs) were proposed to address the issue of expensive computations. These models perform the noising and denoising processes within a reduced latent space, thereby enhancing computational efficiency and improving sampling quality.

\subsection{Face Swapping Models}

     
    \subsubsection*{Structural Guidance Based Models}
    Traditional face-swapping methods, which require manual intervention, benefit greatly from structural information. For faces, landmarks, 3D representations, and segmentation, all of these provide strong structural priors which can be used to generate high-quality swapped images. However, these traditional methods \cite{blanz2004exchanging, nirkin2018face} required manual intervention and could not correctly map the target expressions.
    3D structural priors have recently been combined with GANs for an identity agnostic swapping module \cite{Li_2023_CVPR, jiang2020deeperforensics}. However, these methods are also limited by the accuracy of the underlying 3D models. 

        



    \subsubsection*{Reconstruction Based Models}
    The original deepfakes model \cite{Deepfakes} is based on training two separate autoencoders with a shared encoder and different decoders. However, this approach requires retraining for each unique source-target pair. Conversely, GAN-based methods like SimSwap \cite{chen2020simswap} have been developed to overcome the limitations of identity-specific face swapping, offering a more generalized approach that is not restricted to particular pairs of faces. The method proposed in SimSwap \cite{chen2020simswap} involves segregating the identity data from the decoder component, thereby enabling the entire framework to be universally applicable to any given identity. However, this also generates low-quality results under certain conditions. Typically, subject-agnostic models adjust the intermediate features of the target image to incorporate the identity of the source image. 



\section{Preliminary: Diffusion Models}

\label{sec:ddpm-intro}
    Diffusion Models (DMs) are generative models trained to reverse the earlier added noise using a parameterized Markovian process. Recent studies have demonstrated that DMs are capable of producing images of superior quality \cite{dhariwal2021diffusion, ho2020denoising}. In the following section, we present a concise summary of DMs. 
    
    Starting with any vector $z_0$, the forward noising process produces a series of latents $z_1, ..., z_T$ by adding Gaussian noise by following a variance schedule depicted by $\beta_t \in (0,1)$ at time $t$:
    \begin{equation}
    \begin{aligned}
    \label{eqn:forwardProcess1}
        q(z_t \mid z_{t-1}) &= \mathcal{N}(\sqrt{1-\beta_t} z_{t-1}, \beta_t \mathbf{I})
    \end{aligned}
    \end{equation}
    
    After sufficient noise is added till timestep $T$, the last latent $z_T$ is nearly an isotropic Gaussian distribution. The above equation's closed form can be derived using a simple reparametrization trick. Let $\alpha _t = 1 - \beta _t$ and $\bar{\alpha _t} = \prod_{i=1}^{t} \alpha _i$. Thus, we get the following:

    \begin{equation}
        \begin{aligned}
        \label{eqn:forwardProcessX0}
            q(z_t \mid z_{0}) \sim \mathcal{N}(\sqrt{\bar{\alpha _t}} z_{0}, (1 - \bar{\alpha _t}) \mathbf{I}) \\
            z_t = \sqrt{\bar{\alpha _t}}z_0 + \sqrt{1-\bar{\alpha _t}} \epsilon
        \end{aligned}
    \end{equation}

   Starting from the distribution \( q(z_T) \), a reverse sequence can be generated by sampling the posteriors \( q(z_{t-1}|z_t) \). These posteriors are also Gaussian distributions. To approximate this function, a deep neural network \( p_\theta \) (a 2D U-Net architecture in the context of synthetic image generation) is trained to predict the mean and variance of \( z_{t-1} \) given \( z_t \) as input, or to estimate the noise \( \epsilon_\theta(z_t, t) \), as proposed by Ho et al. \cite{ho2020denoising}. However, vanilla diffusion models are computationally expensive because they operate directly on the pixel space (\( z_t \in \mathbb{R}^{3 \times H \times W} \)). To address this, Rombach et al. \cite{rombach2022high} proposed Latent Diffusion Models (LDMs), which first compress the image to a latent space (\( 64 \times 64 \) in our implementation) and then perform the denoising in this space. Once the denoising process is completed in the latent space, the latent vector is upsampled back to the pixel space. The images are downsampled and upsampled using the encoder and decoder of a pretrained VQGAN \cite{esser2021taming}. Our method uses LDMs along with conditional guidance \cite{dhariwal2021diffusion} utilizing a novel facial guidance module.

    \begin{figure}[t]
          \centering
          \includegraphics[width=\textwidth]{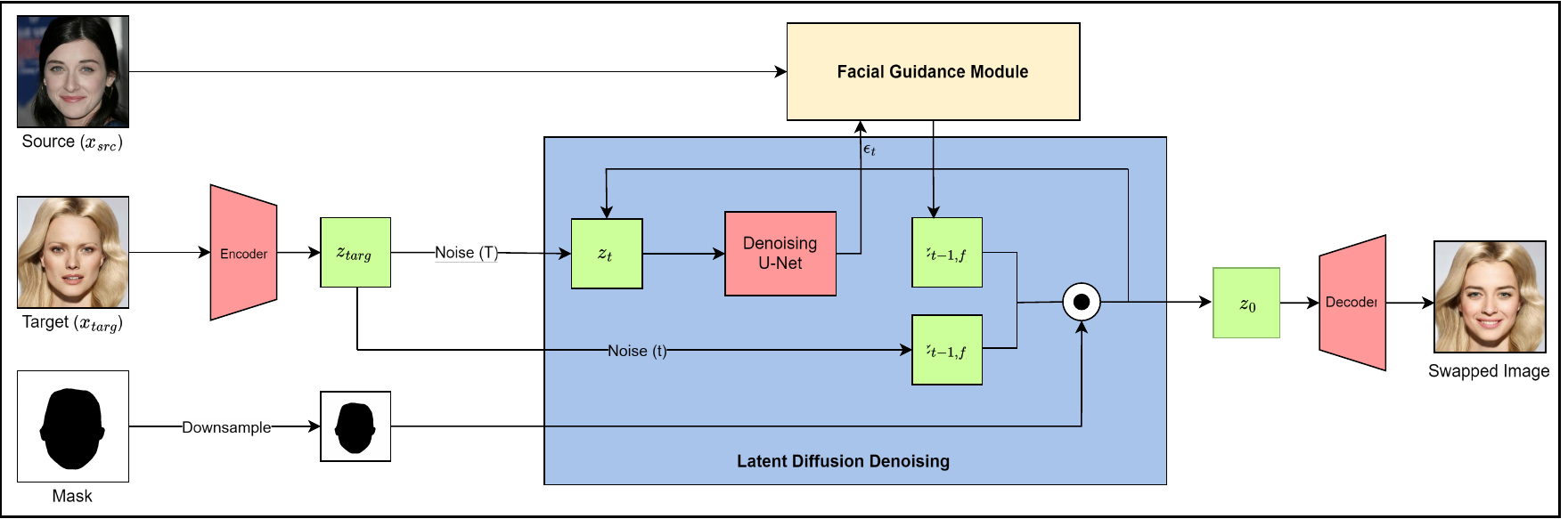}
          \caption{\textbf{Proposed sampling process.} The sampling process begins by encoding the target image into a latent vector using an encoder. The encoder and decoder used in this method come from the same autoencoder based on previous work by Esser et al. \cite{esser2021taming}. Noise is added to this latent vector according to the diffusion noise schedule. Subsequently, a pre-trained U-Net is used to denoise this latent vector. The output of the U-Net is then conditioned using our novel facial guidance module. A downsampled facial mask ensures the masked area acquires the necessary facial characteristics through facial guidance while the background remains constant. Finally, after completing the denoising process, we pass the final latent vector $z_0$ into a decoder to get the swapped image. This entire process is detailed in Algorithm  \ref{algo:LDFaceNet}.}
          \label{fig:sampling-LDFaceNet}
        \end{figure}

\section{Methodology}

        \begin{figure}[h]
          \centering
          \includegraphics[width=12cm]{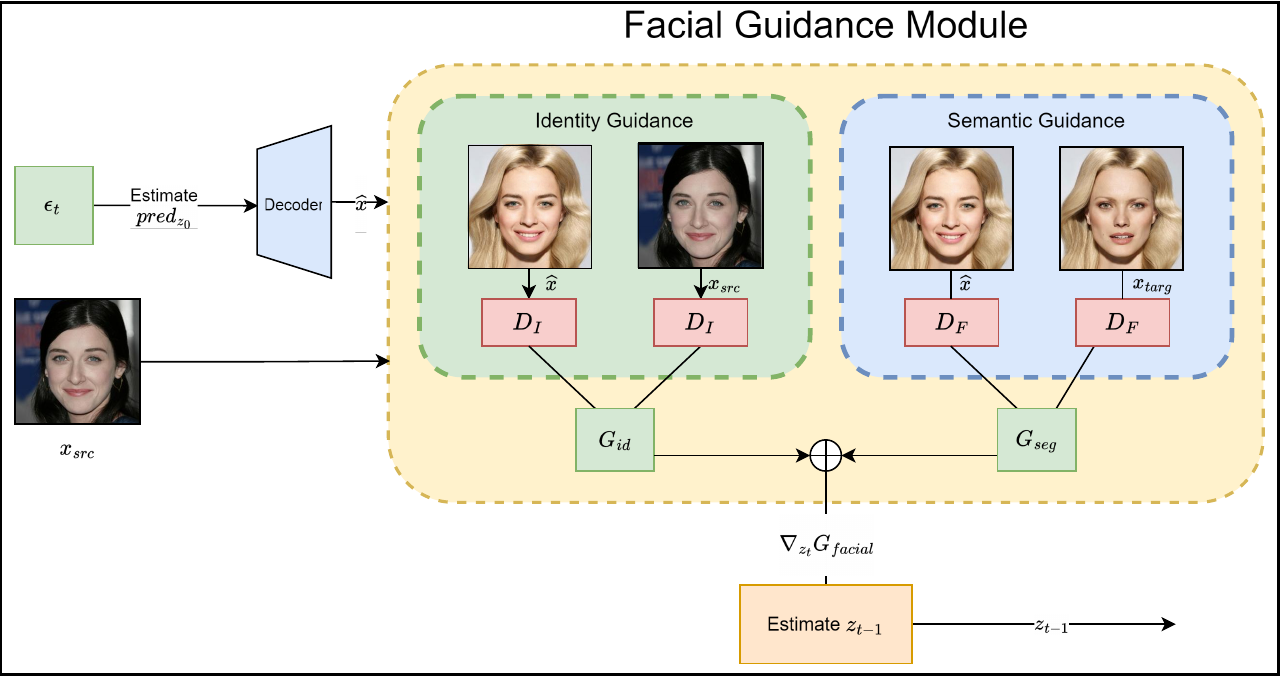}
          \caption{\textbf{Facial Guidance Module.} The latent vector $pred_{z_0}$ is estimated from the output ($\epsilon_t$) of the denoising U-Net. $pred_{z_0}$ is upsampled to get $\widehat{x}$, which approximates what our swapped image would look like after the entire denoising process. $\widehat{x}$ is then used within the identity and segmentation guided modules since these involve using pretrained classifiers trained on normal images and not latent vectors. The embeddings of $\widehat{x}$, $x_{src}$, and $x_{targ}$ are then used as given in Algorithm \ref{algo:LDFaceNet} to calculate facial and segmentation guidance loss modules. These are combined to form the complete facial guidance loss term. The gradient of this facial loss term with respect to $z_t$ is used to guide the reverse diffusion process.}
          \label{fig:ldm-diffusion}
        \end{figure}
    
    In this section, we describe our method, \textsc{LDFaceNet}. Given a source image $x_{src}$, a target image $x_{targ}$, and the facial segmentation mask of $x_{targ}$ as $\mathcal{M}$, our goal is to transfer the facial features of the source image onto the target image while keeping all other attributes of the target image the same. More formally, we need to produce a modified $\widehat{x}$ such that $\widehat{x} \odot \mathcal{M}$ is as similar to $x_{src}$ as possible. Furthermore, $\widehat{x} \odot (1 - \mathcal{M}) \approx x_{targ} \odot (1 - \mathcal{M}) $ to preserve the background and to keep the complementary area nearly the same as before. Here $\odot$ is element-wise multiplication operator.

    In section \ref{id-guided-diffusion}, we extend the latent diffusion approach to support facial editing by incorporating a guiding cosine loss generated by the identity guidance module. Initially, our results indicated that while the swapped images maintained similarity to the source image and preserved the background, they did not map the emotional expressions of the target image.

    Subsequently, in section \ref{seg-guided-diffusion}, we introduce a Euclidean L2 loss term generated by the segmentation guidance module to address this limitation. Our findings demonstrate that our method produces coherent and realistic results. Specifically, the generated images exhibit remarkable similarity to the source image in terms of skin color, eye color, shape, structure, and lighting. Furthermore, they effectively preserve the original facial attributes and emotional expressions of the target image.

    To further validate our approach, we conduct a comprehensive ablation study to evaluate the efficacy of the proposed solutions.
    
    \subsection{Source Identity Guided Diffusion}
    \label{id-guided-diffusion}
    We propose applying facial guidance during the denoising process in order to dictate the facial attributes of generated images. One significant benefit of using this method is that, after training, we can control the image produced by the sampling process's guidance. As a result, we can produce the necessary images without having to retrain the LDM. We use external facial recognition modules to provide guidance in order to take advantage of this advantage. We use the embeddings of these facial recognition modules to calculate the guidance loss term.

    The preferred approach for face recognition uses Deep Convolutional Neural Network (DCNN) embedding to represent faces \cite{taigman2014deepface, schroff2015facenet, sun2014deep,cao2018vggface2, deng2019arcface}. For our experiments, we use a ResNet-50 backbone \cite{he2016deep} pre-trained on the MS1MV3 dataset using the ArcFace \cite{deng2019arcface} loss. The identity guidance module, denoted as $D_I$, constrains the ID vector of $x_{\text{src}}$ to be closer to $\widehat{x}$, the approximation of the swapped image, which is estimated at each denoising step.

    In latent diffusion, the actual denoising occurs in the latent space $Z$. Since we are utilizing pre-trained models trained on actual images, it is necessary to first calculate $\widehat{z}$, an estimation of $z_0$ given $z_t$. Subsequently, we upsample $\widehat{z}$ to obtain an approximation of $\widehat{x}$. This $\widehat{x}$ is then passed as an input into $D_I$ to extract feature embeddings. These embeddings, along with the embeddings of the source image, are processed using a cosine loss.
    
    The overall process is formally described in Algorithm \ref{algo:LDFaceNet} from lines 8 to 10, and the guidance loss is defined as follows:

    \begin{equation}    
        G_{id} = 1 - \cos (D_I (x_{src}), D_I (\widehat{x}))
    \end{equation}

    \subsection{Target Segmentation Guided Diffusion}
    \label{seg-guided-diffusion}
    Using only the Identity Guidance Loss ($G_{id}$) fails to preserve the facial expressions of the target image, such as shape, eye structure, lip structure, and overall facial structure. Consequently, the expressions of $x_{src}$ are mapped onto the generated image, which, even though it gives a satisfactory transfer of identity, but with a loss of the target's expressions.

    \begin{algorithm}
    \caption{\textsc{LDFaceNet} sampling, given a latent diffusion model $\epsilon_\theta(z_t, t)$, Encoder $\varepsilon$, Decoder $\mathcal{D}$, ArcFace Identifer $D_I$ and BiseNet Parser $D_F$}
    \label{algo:LDFaceNet}
    \begin{algorithmic}[1]
    
    \State \textbf{Input}: Source image $x_{src}$, Target image $x_{targ}$, Target mask $\mathcal{M}$, diffusion steps $k$
    \State \textbf{Output}: Face swapped image $\widehat{x_0}$
    
    \State $z_0 = \varepsilon(x_{targ})$
    \State $z_k \gets \text{sample from } \mathcal{N}(\sqrt{\bar{\alpha_k}} z_0, \sqrt{1 - \bar{\alpha_k}} \mathbf{I})$
    \State $m \gets \text{downsampled from } \mathcal{M}$
    
    \For{$t \text{ from } k \text{ to } 1$}
        \State $\epsilon_t = \epsilon_\theta(z_t, t)$
        \State $\widehat{z} = \frac{1}{\sqrt{\bar{\alpha_t}}} \left( z_t - \sqrt{1-\bar{\alpha_t}} \epsilon_t \right)$
        \State $\widehat{x} = \mathcal{D}(\widehat{z})$
        
        \State $G_{id} = 1 - \cos (D_I(x_{src}), D_I(\widehat{x}))$
        \State $G_{seg} = \left\| D_F(x_{targ}) - D_F(\widehat{x}) \right\|_2^2$
        \State $G_{fac} = \lambda_{id}(t) G_{id} + \lambda_{seg}(t) G_{seg}$

        \State $\epsilon_t = \epsilon_t + \sqrt{1 - \bar{\alpha_t}} \nabla_{z_t} G_{fac}$         \Comment{guide $\epsilon_t$ using the gradient of $G_{fac}$}
        \State $\widehat{z} = \frac{1}{\sqrt{\bar{\alpha_t}}} \left( z_t - \sqrt{1-\bar{\alpha_t}} \epsilon_t \right)$
        
        \State $z_{t-1, fg} \gets \text{sample from } \mathcal{N}\left(\sqrt{\bar{\alpha_{t-1}}} \widehat{z} + \sqrt{(1 - \bar{\alpha_{t-1}} - \Sigma^2)} \epsilon_t , \Sigma\right)$
        \State $z_{t-1, bg} \gets \text{sample from } \mathcal{N}(\sqrt{\bar{\alpha_k}} z_0, \sqrt{1 - \bar{\alpha_k}} \mathbf{I})$
        \State $z_{t-1} = z_{t-1, fg} \odot m + z_{t-1, bg} \odot (1-m)$
    \EndFor
    
    \State $\widehat{x_0} = \mathcal{D}(z_0)$

    \State $\textbf{return } \widehat{x_0}$

    \end{algorithmic}
\end{algorithm}

    To address this issue, we use BiseNet \cite{yu2018bisenet} as a face segmentation model ($D_F$), which predicts pixel-wise probabilities for facial components (such as the nose, eyebrows, and eyes). This allows us to explicitly match the facial expressions of the synthesized image to those of the target. Through segmentation guidance, the generated image retains similarity to the target image in terms of expression, pose, and shape. The Euclidean L2 distance between the two segmentation maps is then calculated. The formal segmentation guidance loss term is defined as follows:
    
    \begin{equation}
        G_{seg} = || D_F (x_{targ}) - D_F (\widehat{x}) ||_2 ^2
    \end{equation}

    The final facial guidance loss term ($G_{fac}$) is calculated by combining the identity guidance loss ($G_{id}$) and the segmentation loss ($G_{seg}$), each weighted by their respective lambdas. These lambdas are not constants. We discovered that using a decreasing step function for these lambdas as denoising progresses performs better than keeping them constant. In our experiment, we decrease the lambdas according to a stepwise decreasing schedule. $G_{fac}$ is formally defined as follows:
    \begin{equation}
        G_{fac} = \lambda_{id}(t) G_{id} + \lambda_{seg}(t) G_{seg}
    \end{equation}

    \subsection{Background Preservation}
    Burt and Adelson \cite{burt1987laplacian} demonstrated that images can be effectively blended by combining each level of their Laplacian pyramids separately. Building on this method, we blend at different timesteps at the latent level to introduce varying amounts of noise as the denoising process progresses. The rationale is that, during each step of this sampling process, we superimpose noisy latents in the form of the background onto a set of naturally noisy images. Directly merging two noisy images from the same timestep often results in incoherence due to differing distributions. However, the subsequent diffusion step projects the result onto the manifold of the next level, enhancing coherence.
    
    Formally, starting with a noisy latent $z_t$, we execute a guided diffusion step that produces a latent $z_{t-1, fg}$. Concurrently, we obtain $z_{t-1, bg}$ using Equation (\ref{eqn:forwardProcessX0}). These two latents are then blended using a mask $m$, which is downsampled from the original target mask $\mathcal{M}$:
    
    \begin{equation}
    z_{t-1} = z_{t-1, fg} \odot m + z_{t-1, bg} \odot (1-m)
    \end{equation}


\section{Results and Discussion}
\label{sec:results}
    In this section, we provide a comprehensive analysis of the \textsc{LDFaceNet} model. We assess the proposed approach from both a quantitative and qualitative perspective to ascertain its robustness. In addition, we perform a few ablation experiments to evaluate the relative contributions of different components of the model, which highlight the importance of their presence.

    \begin{figure}
    \centering    
    \resizebox{.9\textwidth}{!}{
        \begin{tabular}{@{}c@{}c@{}c@{}c@{}c@{}c@{}c@{}}
            \toprule
            Source & Target & E4S \cite{liu2023e4s} & DiffFace \cite{kim2022diffface} & MegaFS \cite{zhu2021one} & MobileFace \cite{xu2022mobilefaceswap} & Ours\\
            \midrule

            \includegraphics[width=2.2cm]{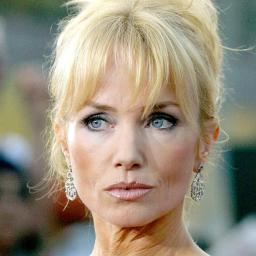} & \includegraphics[width=2.2cm]{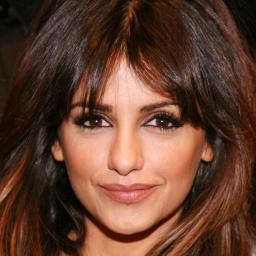} & \includegraphics[width=2.2cm]{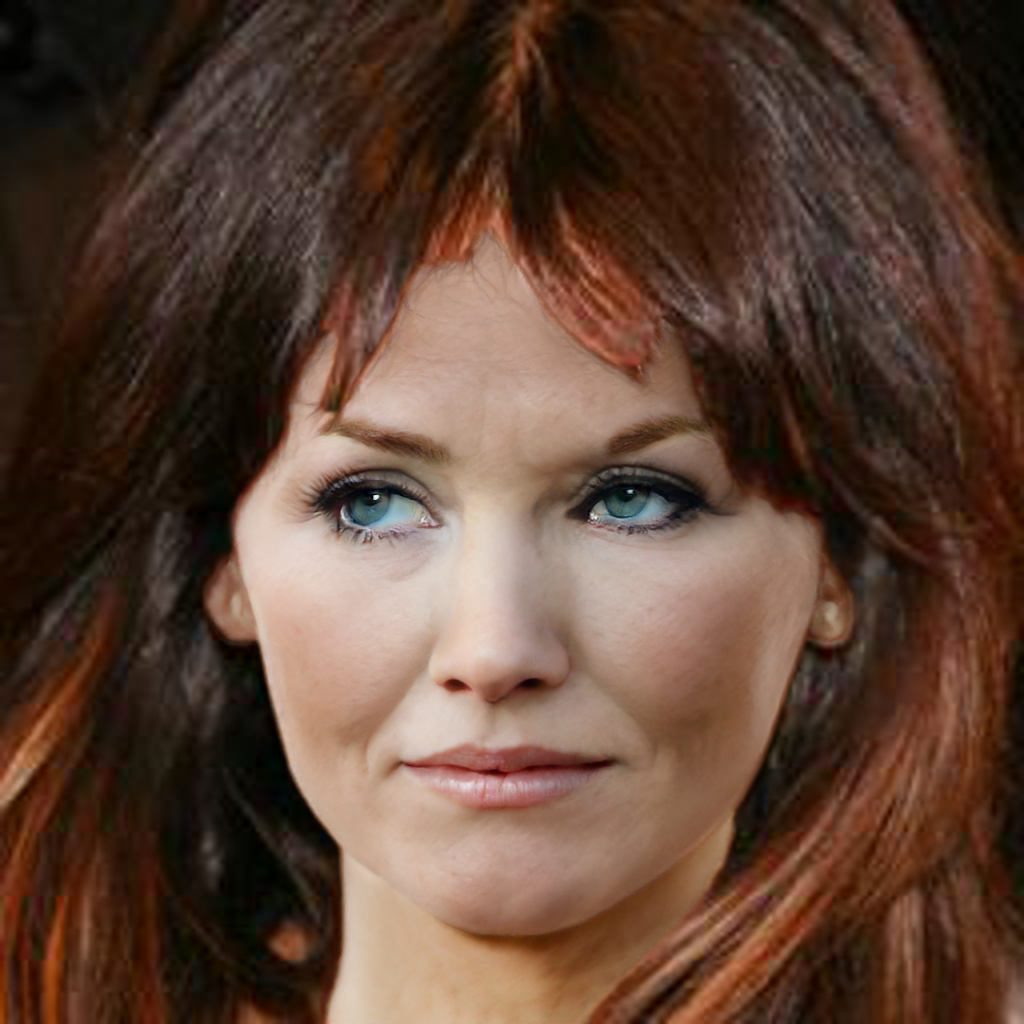} & \includegraphics[width=2.2cm]{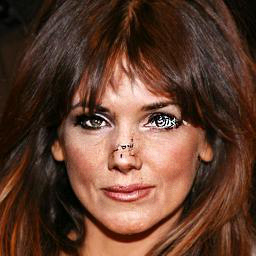} & \includegraphics[width=2.2cm]{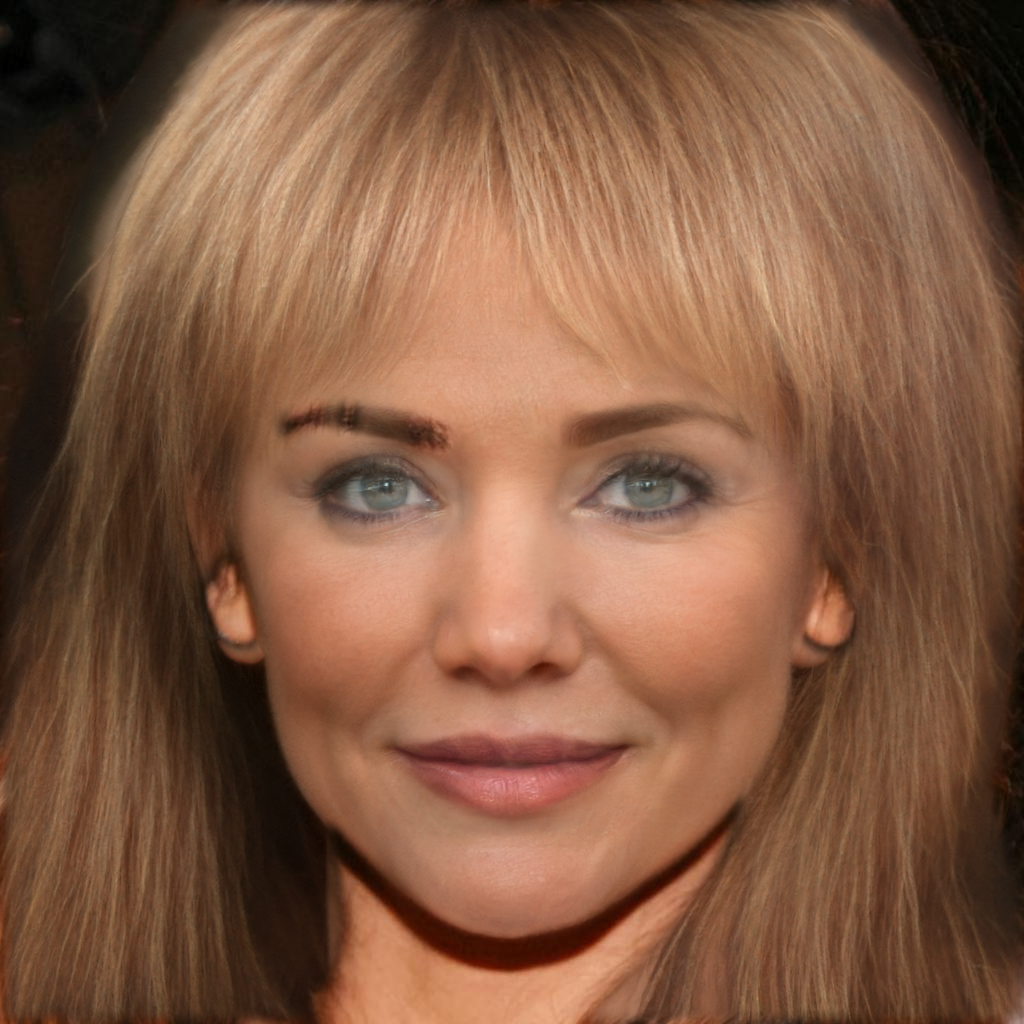} & \includegraphics[width=2.2cm]{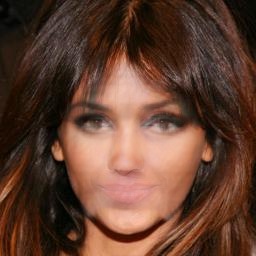} & \includegraphics[width=2.2cm]{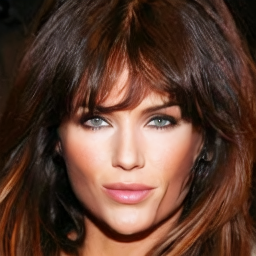} \\

            \includegraphics[width=2.2cm]{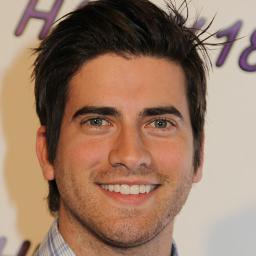} & \includegraphics[width=2.2cm]{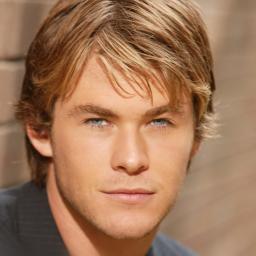} & \includegraphics[width=2.2cm]{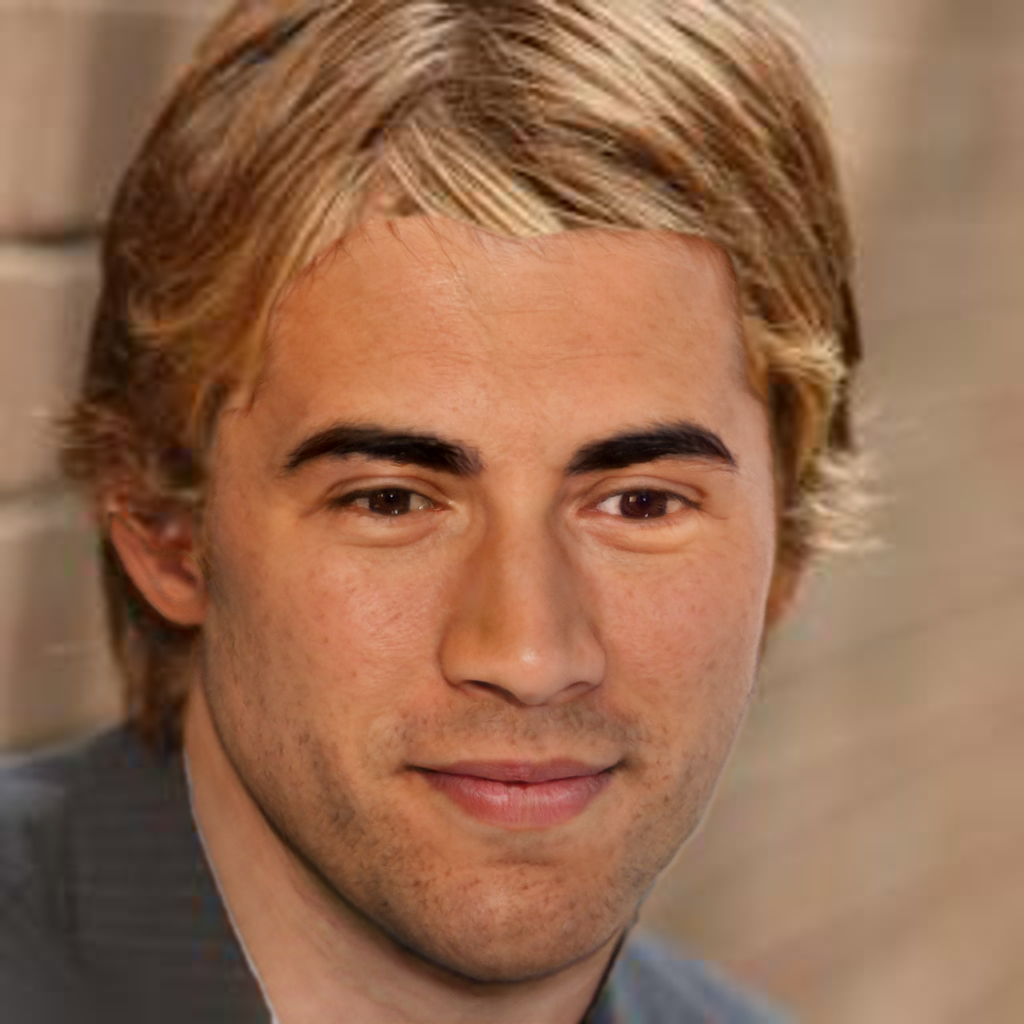} & \includegraphics[width=2.2cm]{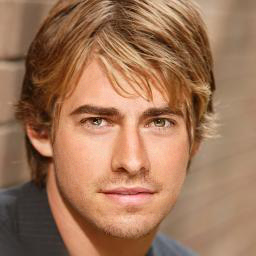} & \includegraphics[width=2.2cm]{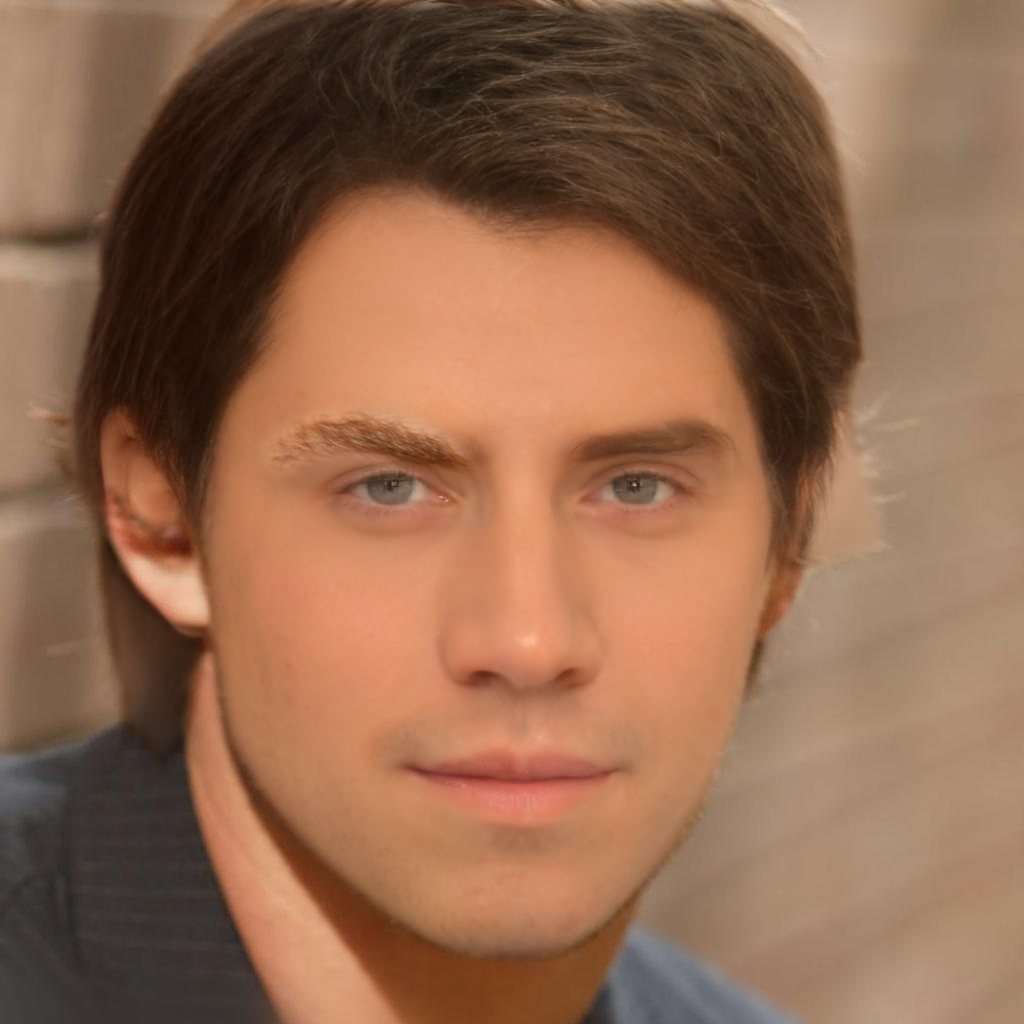} &
            \includegraphics[width=2.2cm]{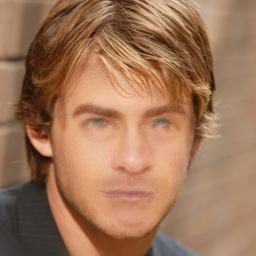} & 
            \includegraphics[width=2.2cm]{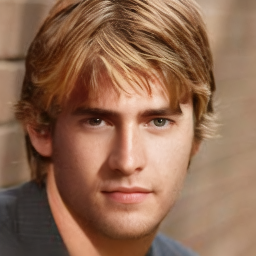} \\

            \includegraphics[width=2.2cm]{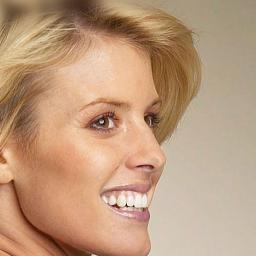} & \includegraphics[width=2.2cm]{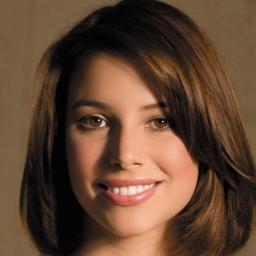} & \includegraphics[width=2.2cm]{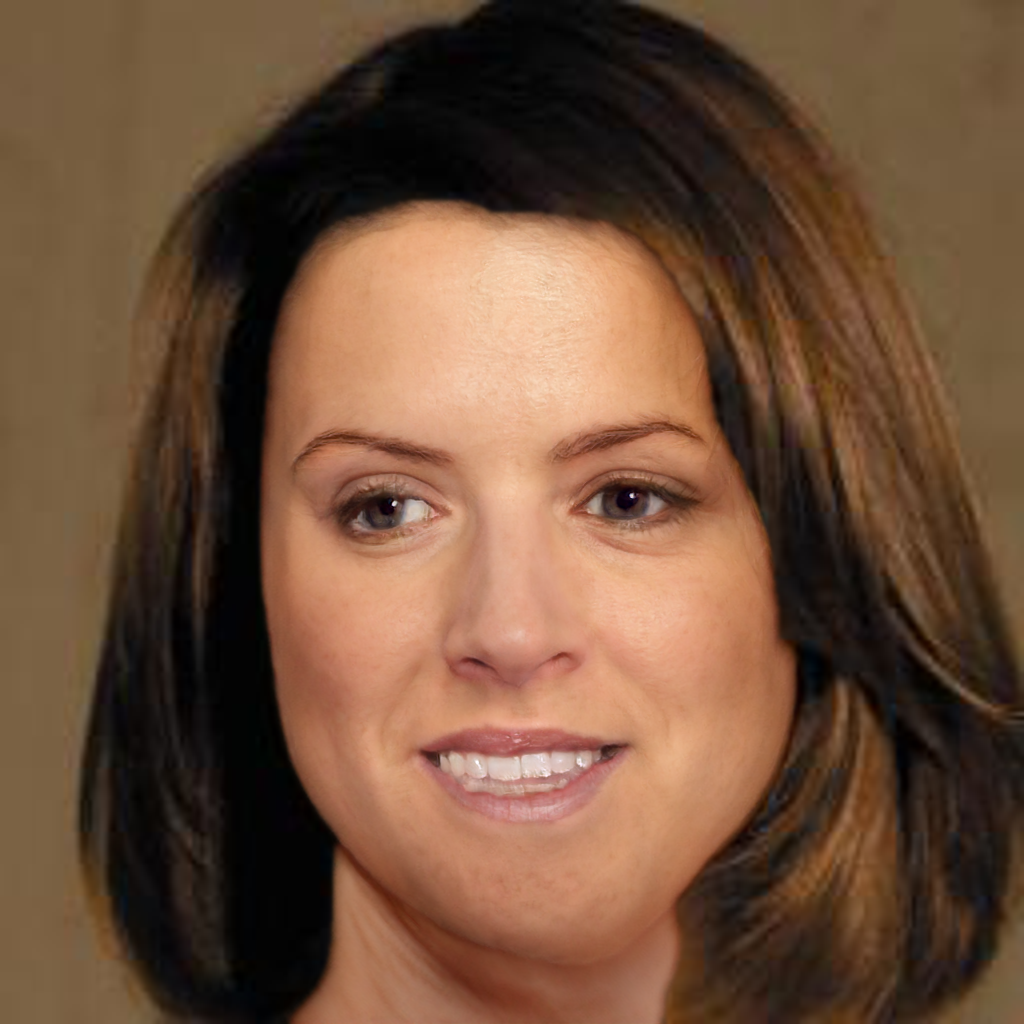} & \includegraphics[width=2.2cm]{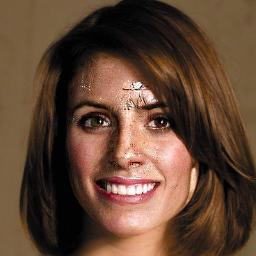} & \includegraphics[width=2.2cm]{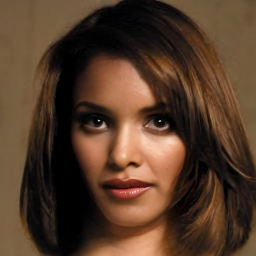} & \includegraphics[width=2.2cm]{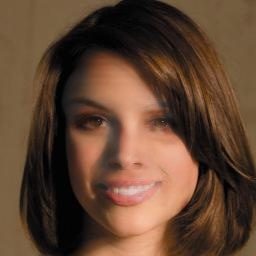} & \includegraphics[width=2.2cm]{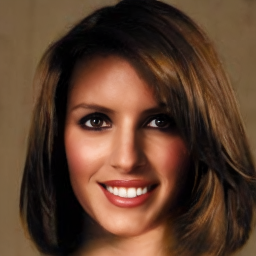} \\

            \includegraphics[width=2.2cm]{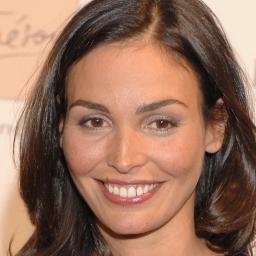} & \includegraphics[width=2.2cm]{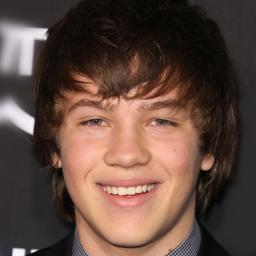} & \includegraphics[width=2.2cm]{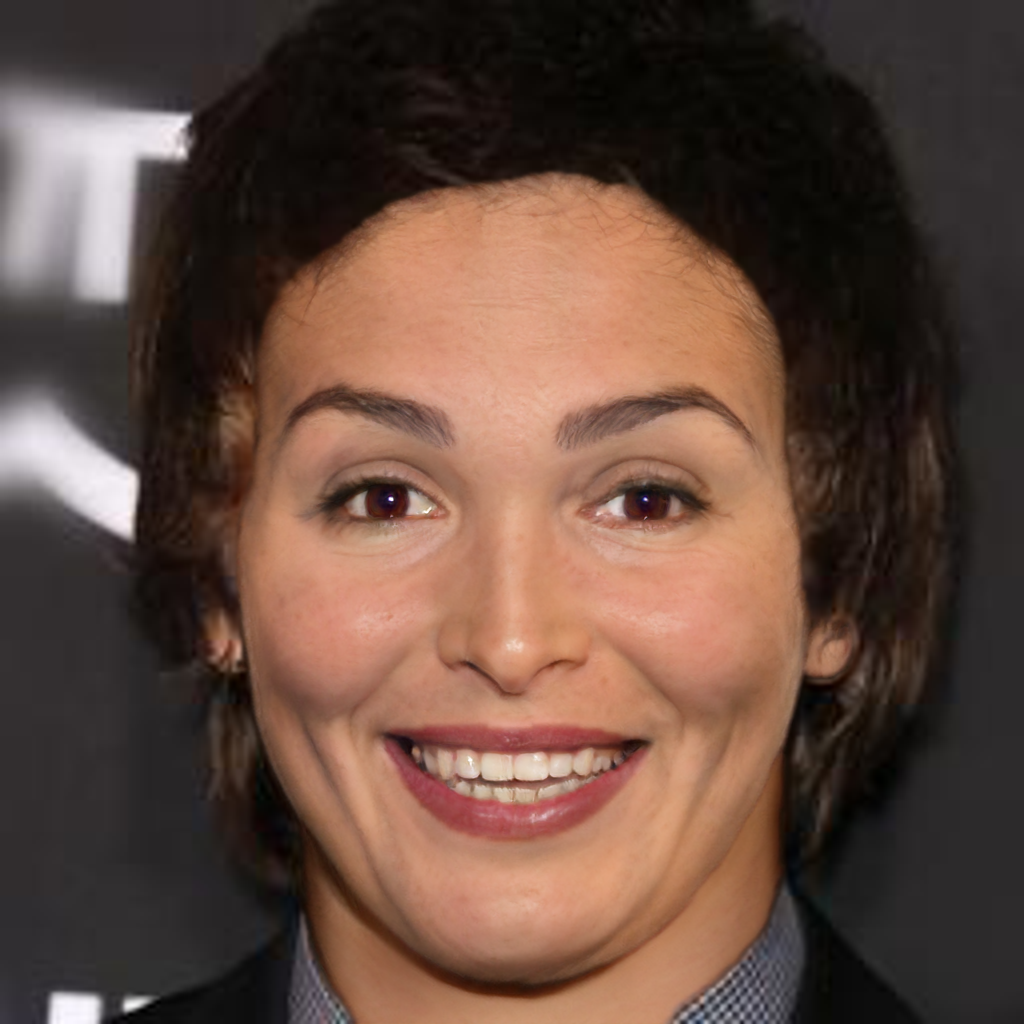} & \includegraphics[width=2.2cm]{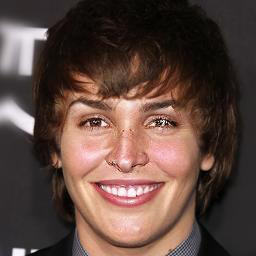} & \includegraphics[width=2.2cm]{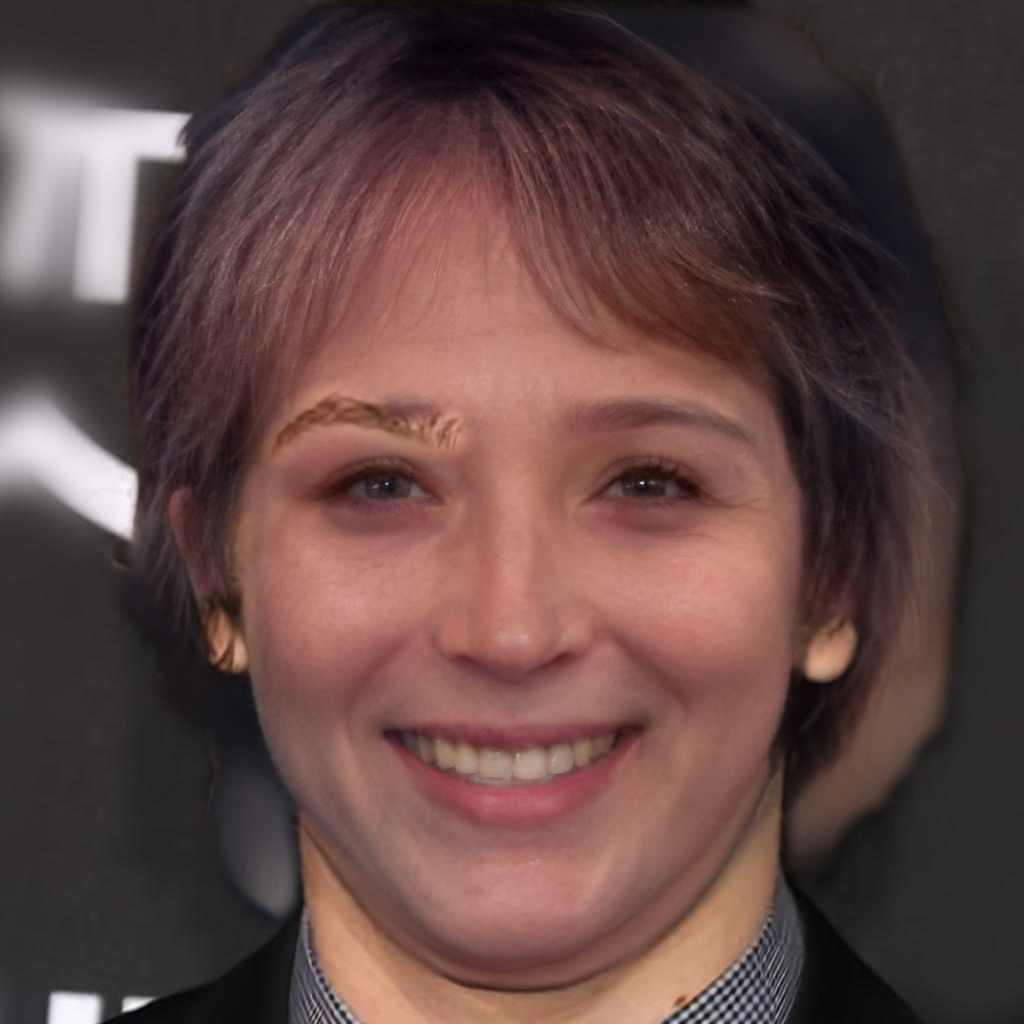} & \includegraphics[width=2.2cm]{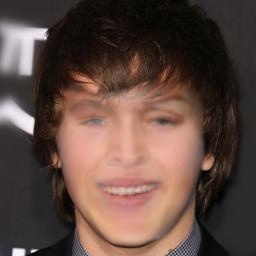} & \includegraphics[width=2.2cm]{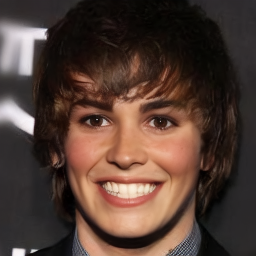} \\

            \includegraphics[width=2.2cm]{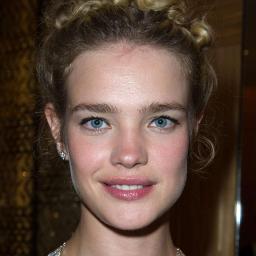} & \includegraphics[width=2.2cm]{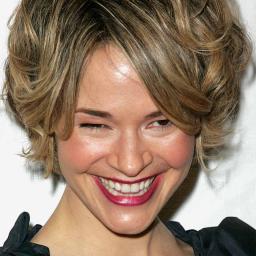} & \includegraphics[width=2.2cm]{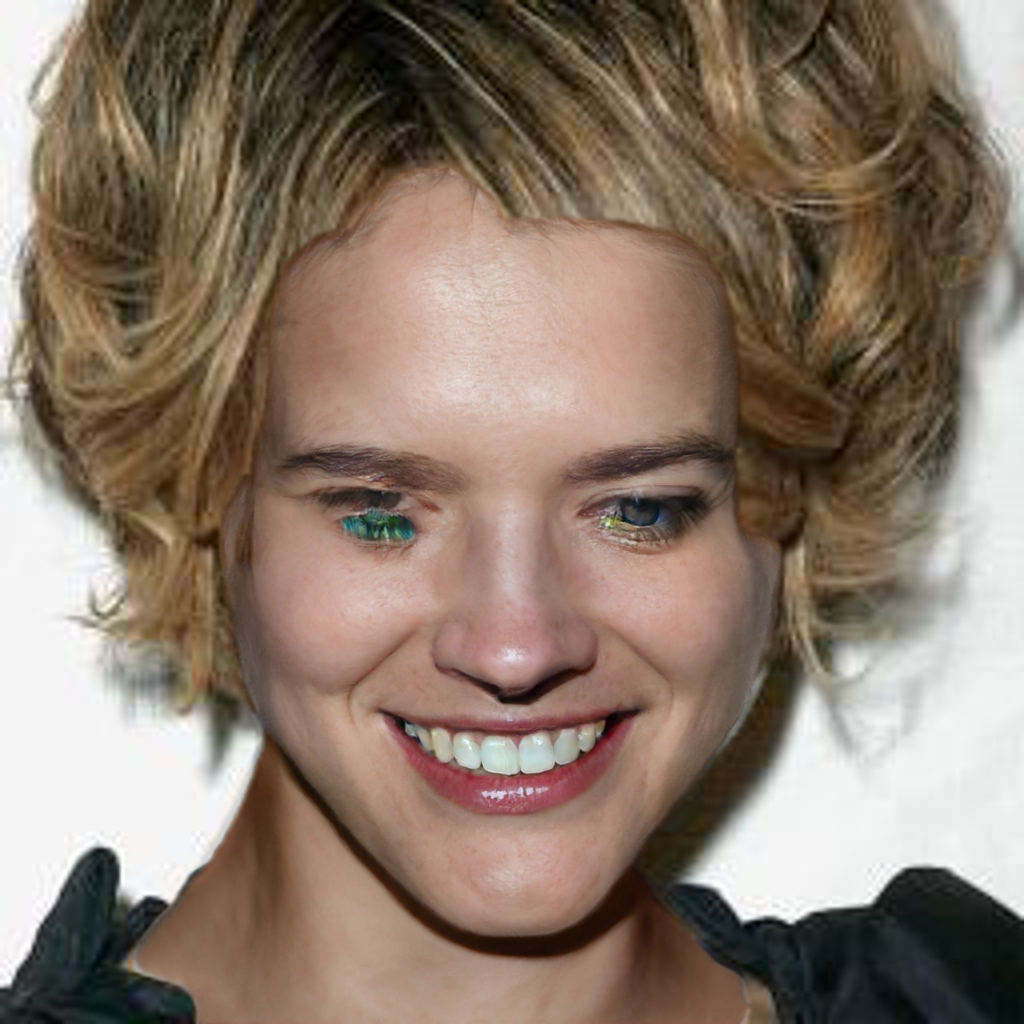} & \includegraphics[width=2.2cm]{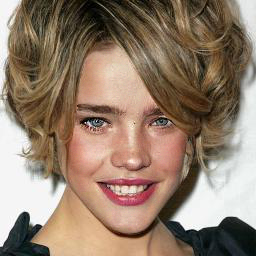} & \includegraphics[width=2.2cm]{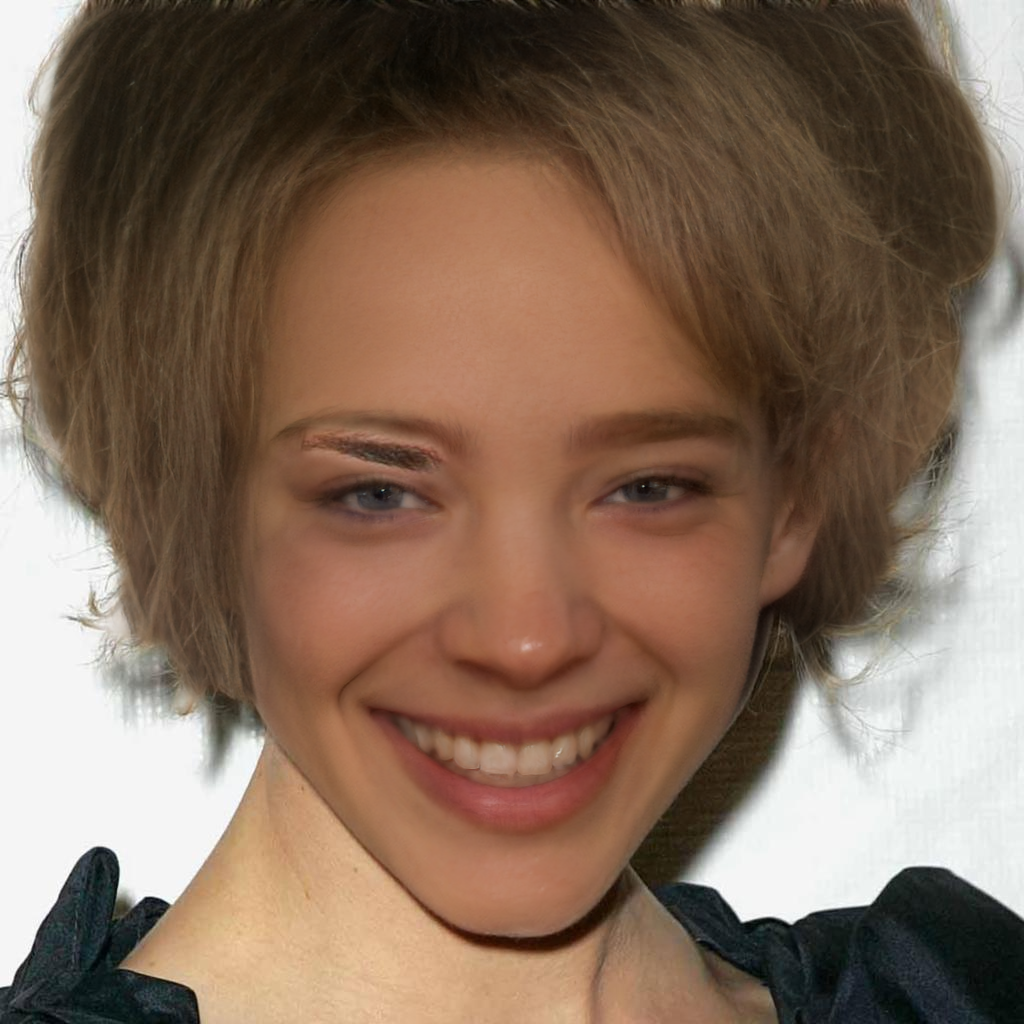} & \includegraphics[width=2.2cm]{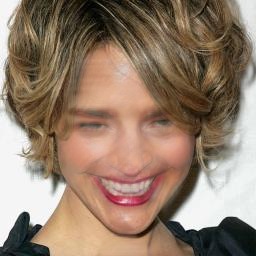} & \includegraphics[width=2.2cm]{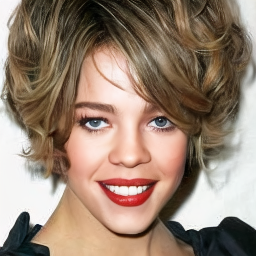} \\
            
            \bottomrule
        \end{tabular}
    }
    \caption{\textbf{Qualitative Results.} Our method achieves high-fidelity results, better preserving source identity and target facial attributes than other methods. It also handles occlusions and partial views robustly.}
    \label{fig:gen_result}
\end{figure}

\begin{table}[h]
    \centering
    \begin{threeparttable}
        \caption{Comparative results of the \textsc{LDFaceNet} and other existing face swapping methods over CelebA dataset \cite{liu2018largeCelebA}.}
        \vspace{0.1in}
        \begin{tabular}{l|l|l|l}
            \hline
            Method & ID similarity $\uparrow$ & Pose $\downarrow$ & Expr. $\downarrow$  \\ \hline
            MobileFace (AAAI'22) \cite{xu2022mobilefaceswap} & 0.25 & 2.52 & 3.72 \\
            MegaFS (CVPR'21) \cite{zhu2021one} & 0.26 & 2.48 & 3.27 \\
            DiffFace \cite{kim2022diffface} & 0.55 & 2.40 & 2.71 \\
            E4S (CVPR'23) \cite{liu2023e4s} & 0.61 & 2.31 & 2.80 \\
            \textbf{\textsc{LDFaceNet}} & \textbf{0.67} & \textbf{2.18} & \textbf{2.55} \\ \hline
        \end{tabular}
    \end{threeparttable}
    \label{tab:quantitative_results}
\end{table}

    \subsection{Quantitative and Qualitative results}
    To generate the results, we use \textsc{LDFaceNet} with the pre-trained LDM model, ArcFace identity extractor \cite{deng2019arcface}, and BiseNet face parser \cite{yu2018bisenet}. The generated images are obtained through the sampling process detailed in Algorithm \ref{algo:LDFaceNet}. For quantitative analysis we use three metrics. The ability to transfer structural attributes is indicated by the pose error and expression error. These errors are represented as L2 distances between the pose and expression feature vectors of the swapped image and target image. Pose and expression vectors are generated using pre-trained estimators, specifically Hopenet \cite{ruiz2018fine} and a 3D face reconstruction model \cite{deng2019accurate}, respectively. We also calculate the ID similarity score, which is the cosine similarity between swapped faces and their corresponding sources. 
    
    We present the results side-by-side for each pair of source and target images in figure \ref{fig:gen_result}. It clearly demonstrates the ability of \textsc{LDFaceNet} to generate realistic images by transferring the facial features and expressions of the target image onto the source image. The generated images are compared with other state-of-the-art methods for a thorough analysis. Further figure \ref{fig:gen_result} analyses the quantitative performance by showing the cosine similarity (higher the better), pose error (lower the better) and expression error (lower the better). The numbers also unequivocally demonstrate our model's superior performance compared to other recent face-swapping models. It is evident that \textsc{LDFaceNet} outperforms the previous techniques, including recent models like E4S (CVPR'23) by a considerable margin. 
    
    Overall, the results demonstrate that \textsc{LDFaceNet} can produce high-quality images that closely resemble the source image while retaining the characteristics of the target image. The generated images show realistic facial expressions, lighting, and background, which are crucial for creating realistic face swaps. These results highlight the potential of \textsc{LDFaceNet} as a powerful tool for image manipulation and face swapping.

    \subsection{Ablation Study}
\begin{figure}
    \centering
    \includegraphics[width=0.75\textwidth]{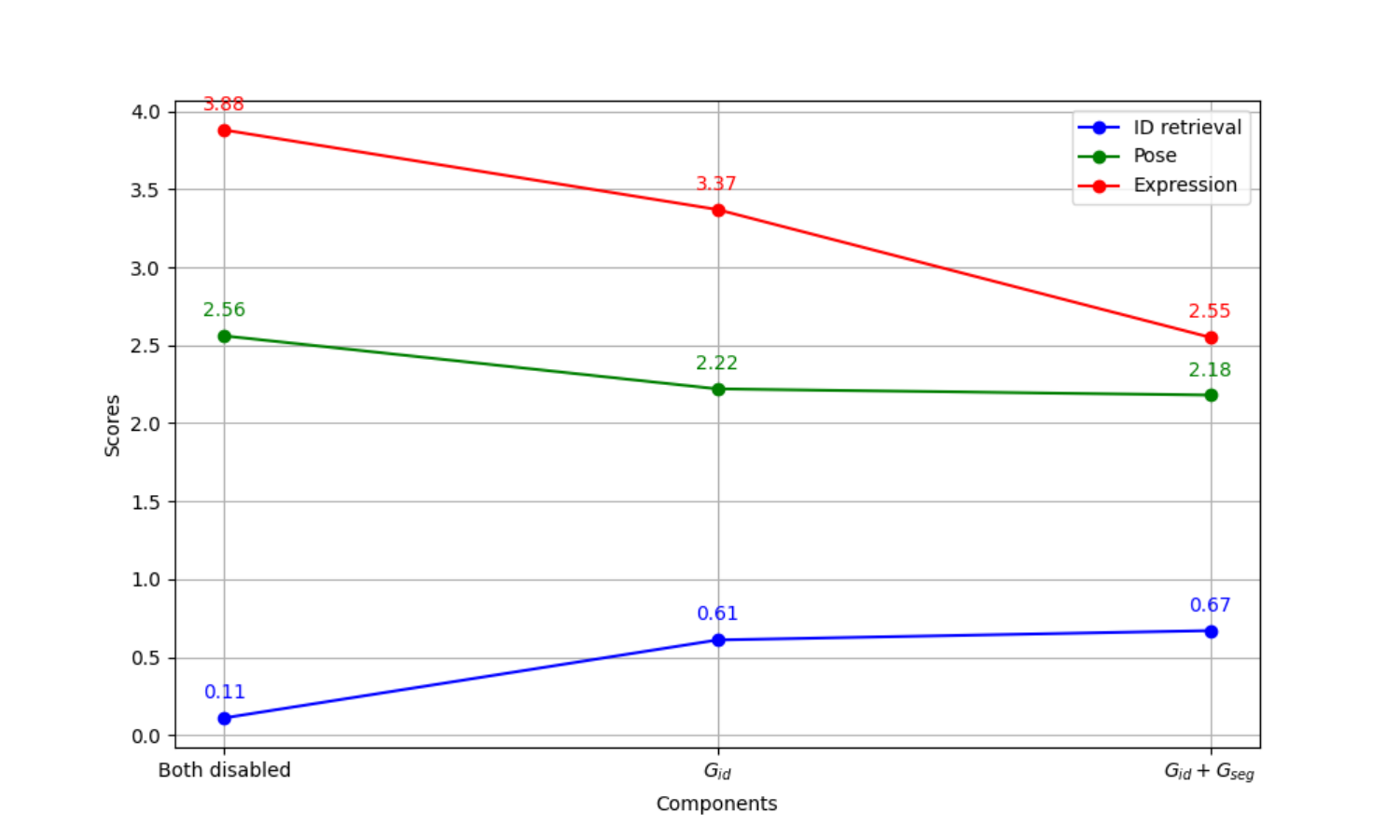}
    \caption{Comparative performance of ablation experiments. x-axis represents the three variants of \textsc{LDFaceNet}. The three lines describe the performance of each variant for three metrics.} 
    \label{fig:graph-ablation}
\end{figure}

\begin{figure}
    \centering
    
    \resizebox{\textwidth}{!}{
        \begin{tabular}{@{}c@{}c@{}c@{}c@{}c@{}}
            \toprule
            Source & Target & Both Disabled & $G_{id}$ only & $G_{id} + G_{seg}$ \\
            \midrule
            \includegraphics[width=3cm]{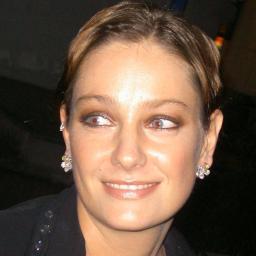} & \includegraphics[width=3cm]{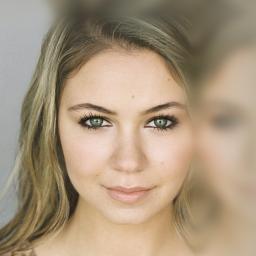} & \includegraphics[width=3cm]{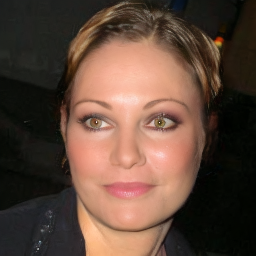} & \includegraphics[width=3cm]{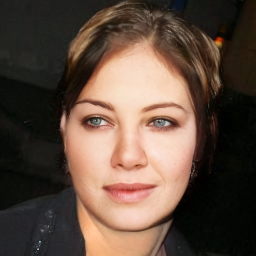} & \includegraphics[width=3cm]{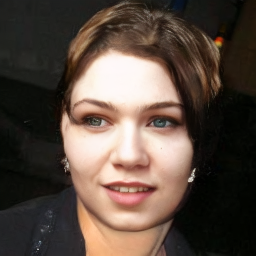} \\
            \includegraphics[width=3cm]{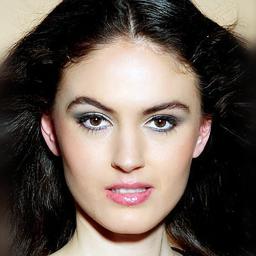} & \includegraphics[width=3cm]{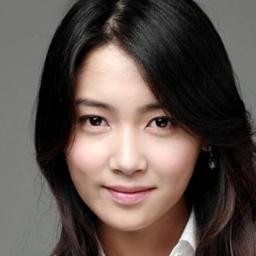} & \includegraphics[width=3cm]{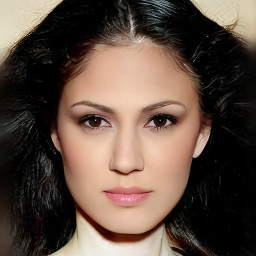} & \includegraphics[width=3cm]{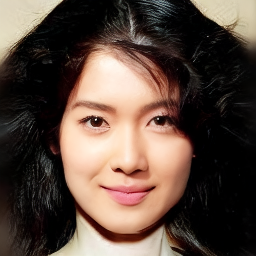} & \includegraphics[width=3cm]{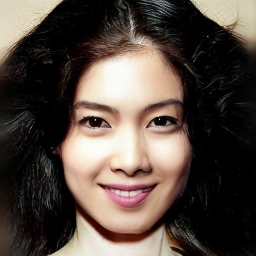} \\
            \bottomrule
        \end{tabular}
    }
    \caption{\textbf{Ablation study generated examples.} Qualitative results of the ablation study demonstrate the importance of the individual guidance modules within our comprehensive facial guidance module. Disabling both modules causes the model to behave like a simple LDM, attempting to reconstruct the target image as it is. When only the identity guidance is enabled, the identity of the target is mapped, but the source's facial expressions are lost. Further, when both modules are enabled, the model successfully preserves both the source's identity and the target's facial expressions.}
    \label{fig:ablations}
\end{figure}

    To assess the significance of the identity and segmentation guidance modules, we conducted experiments with three different configurations: disabling only the segmentation module, disabling both modules, and enabling both modules. The results of these experiments are shown in Figure \ref{fig:ablations}, highlighting the importance of both modules. Specifically, when the segmentation module is disabled, the source's facial expression is copied onto the result image, and the target's facial expression is lost. This demonstrates the essential role of the segmentation module in preventing the loss of target's facial expression.When both modules are disabled, our model attempts to reconstruct the source image without any guidance. As a result, the generated image appears visually similar to the target image, with no discernible change in facial features or attributes. This highlights the crucial role played by the identity and segmentation guidance modules in achieving facial swapping with controllable and desirable results. We present these obeservations and quantitative scores through Figure \ref{fig:graph-ablation} for a better visualization of contribution of the two components. 


    Our ablation experiments indicate that the identity and segmentation guidance modules are critical for achieving high-quality facial swapping with \textsc{LDFaceNet}. By incorporating facial guidance, we achieve better results in visual fidelity and attribute preservation. Additionally, our model can apply different levels of guidance to balance identity and attribute preservation. While \textsc{LDFaceNet} performs excellently, there are opportunities for further enhancement. One direction is to train a new diffusion model on CelebA using classifier-free guidance \cite{ho2022classifier}. Incorporating more identity and face-parser networks into an ensemble could also create a more robust guidance loss function, further refining our model's capability for face swaps.


\section{Conclusion}
\textsc{LDFaceNet} is a guided diffusion model for facial swapping that leverages facial segmentation and facial recognition modules for a conditioned denoising process. With its unique guidance loss functions, \textsc{LDFaceNet} offers directional guidance to the diffusion process, and can incorporate supplementary facial guidance for desired outcomes without retraining. \textsc{LDFaceNet} improves upon previous GAN-based approaches by utilizing the potential of the diffusion model for facial swapping, resulting in superior visual outcomes and greater diversity.  

In conclusion, \textsc{LDFaceNet} offers a promising new approach to facial swapping by utilizing guided diffusion, segmentation, and recognition modules. The results demonstrate the proposed method's efficacy and highlight the diffusion model's potential for face swapping tasks. This study represents a significant contribution to the field of face swapping using diffusion models and serves as a foundation for future research in this area.


\bibliographystyle{unsrt}  
\bibliography{references}

\end{document}